\documentclass[conference]{IEEEtran}
\IEEEoverridecommandlockouts
\usepackage{cite}
\usepackage{amsmath,amssymb,amsfonts}
\usepackage{algorithmic}
\usepackage{graphicx}
\usepackage{textcomp}
\usepackage{xcolor}
\usepackage{booktabs}
\usepackage{stfloats}
\usepackage{enumitem}
\usepackage{tikz}
\usepackage[colorlinks=true,linkcolor=blue,urlcolor=blue]{hyperref}

\def\BibTeX{{\rm B\kern-.05em{\sc i\kern-.025em b}\kern-.08em
    T\kern-.1667em\lower.7ex\hbox{E}\kern-.125emX}}

\begin{document}

\title{VLAD: A VLM-Augmented Autonomous Driving Framework with Hierarchical Planning and Interpretable Decision Process\\
}

\author{
\IEEEauthorblockN{1\textsuperscript{st} Cristian Gariboldi\textsuperscript{*}}
\IEEEauthorblockA{\textit{Gifu University} \\
Gifu, Japan}
\and
\IEEEauthorblockN{2\textsuperscript{nd} Hayato Tokida}
\IEEEauthorblockA{\textit{Gifu University} \\
Gifu, Japan}
\and
\IEEEauthorblockN{3\textsuperscript{rd} Ken Kinjo}
\IEEEauthorblockA{\textit{DENSO CORPORATION} \\
Tokyo, Japan}
\and
\IEEEauthorblockN{4\textsuperscript{th} Yuki Asada}
\IEEEauthorblockA{\textit{DENSO CORPORATION} \\
Tokyo, Japan}
\and
\IEEEauthorblockN{5\textsuperscript{th} Alexander Carballo}
\IEEEauthorblockA{\textit{Gifu University} \\
Gifu, Japan}
}

\newcommand\copyrighttext{%
  \footnotesize \textcopyright 2025 IEEE. Personal use of this material is permitted.
  Permission from IEEE must be obtained for all other uses, in any current or future
  media, including reprinting/republishing this material for advertising or promotional
  purposes, creating new collective works, for resale or redistribution to servers or
  lists, or reuse of any copyrighted component of this work in other works.
  DOI: to be added upon publication.}
\newcommand\copyrightnotice{%
\begin{tikzpicture}[remember picture,overlay]
\node[anchor=south,yshift=10pt] at (current page.south) {\fbox{\parbox{\dimexpr\textwidth-\fboxsep-\fboxrule\relax}{\copyrighttext}}};
\end{tikzpicture}%
}

\maketitle

\copyrightnotice
\begin{abstract}
Recent advancements in open-source Visual Language Models (VLMs) such as LLaVA, Qwen-VL, and Llama have catalyzed extensive research on their integration with diverse systems. The internet-scale general knowledge encapsulated within these models presents significant opportunities for enhancing autonomous driving perception, prediction, and planning capabilities. In this paper we propose VLAD, a vision-language autonomous driving model, which integrates a fine-tuned VLM with VAD, a state-of-the-art end-to-end system. We implement a specialized fine-tuning approach using custom question-answer datasets designed specifically to improve the spatial reasoning capabilities of the model. The enhanced VLM generates high-level navigational commands that VAD subsequently processes to guide vehicle operation. Additionally, our system produces interpretable natural language explanations of driving decisions, thereby increasing transparency and trustworthiness of the traditionally black-box end-to-end architecture. Comprehensive evaluation on the real-world nuScenes dataset demonstrates that our integrated system reduces average collision rates by 31.82\% compared to baseline methodologies, establishing a new benchmark for VLM-augmented autonomous driving systems.
\end{abstract}


\section{Introduction}
End-to-end autonomous driving architectures have attracted significant attention in recent years due to their inherent advantages over traditional modular systems, particularly their ability to mitigate cascading errors that can accumulate and propagate over the different components. These data-driven approaches offer superior scalability and eliminate the need for hand-engineered rules or complex cost functions typically required by optimization-based models. However, the black-box nature of end-to-end architectures presents substantial challenges for interpretability and reliability, crucial considerations when deploying such systems in safety-critical applications like autonomous vehicles.
Visual Language Models (VLMs) have emerged as powerful tools for enhancing end-to-end autonomous driving systems, leveraging their extensive world knowledge and remarkable generalization to unseen prompts. Their natural language processing abilities are particularly valuable for generating comprehensible explanations of driving decisions, thereby addressing the critical needs for transparency and reliability in autonomous navigation systems.

In this paper, we introduce VLAD (Vision Language Autonomous Driving), a novel hybrid framework that integrates a specialized Visual Language Model with a state-of-the-art end-to-end autonomous driving system, VAD\cite{b1}. Building upon methodological approaches similar to those found in Senna\cite{b2}, we enhance the spatial reasoning and driving decision capabilities of our VLM through a carefully curated question-answer dataset. This dataset is generated using a teacher model (LLaVA-v1.6-34b\cite{b3}) and includes comprehensive environmental perception data from camera inputs, motion predictions for surrounding traffic participants (with a particular focus on vulnerable road users such as pedestrians and cyclists), and high-level planning strategies.
To balance planning accuracy with computational efficiency, we fine-tune a more lightweight model (Vicuna-7b-v1.5\cite{b4}) on this specialized dataset. This approach ensures both precise meta-action selection and real-time inference capabilities for generating navigational commands and accompanying explanations. The meta-actions selected by the VLM, based on its visual understanding of the driving environment, are subsequently processed by VAD's planning component to generate executable trajectories.

The contribution of our paper is threefold:
\begin{enumerate}[label=(\roman*)]
\item A distinctive feature of our system is its capacity to provide real-time natural language explanations for each planning decision, ensuring complete transparency and justification for navigational choices based on comprehensive scene analysis.
\item Our evaluation on the nuScenes\cite{b5} dataset demonstrates that VLAD significantly outperforms existing state-of-the-art baselines, particularly in safety-critical metrics such as collision rates. The results confirm that VLMs, when enhanced with domain-specific driving knowledge, can substantially improve safety in autonomous driving trajectory planning through their sophisticated scene understanding capabilities.
\item To the best of our knowledge, VLAD represents the first framework that utilizes fine-tuned VLMs for both hierarchical planning supervision and interpretable decision processes integrated with end-to-end autonomous driving systems, establishing a new paradigm for explainable autonomous navigation.

\end{enumerate}

\section{Related Work}

\subsection{VLMs as End-to-End Autonomous Planners}

Previous works have already put effort into integrating Large Language Models (LLMs) and Visual Language Models (VLMs) for autonomous driving. LanguageMPC\cite{b6} takes advantage of LLM high-level decision-making capabilities to tune the parameters of an MPC, influencing its decision-making and driving style. DriveGPT4\cite{b7} processes multi-frame video inputs and textual queries to predict vehicle control signals, interpret actions, provide reasoning, and address user questions.
LingoQA\cite{b8}, DriveLM\cite{b9} and DriveBench\cite{b10} provide specific question-answer datasets and benchmarks for fine-tuning and evaluating VLMs perception, prediction and planning capabilities in autonomous driving settings. Many planners integrated with VLMs have been designed and tested in closed-loop settings (like CARLA\cite{b11} simulator). DriveAnywhere\cite{b12}, LMDrive\cite{b13}, CarLLaVA\cite{b14}, DriveMLM\cite{b15}, DriveVLM-Dual\cite{b16}, VLM-RL\cite{b17}, SimLingo\cite{b18} and Pix2Planning\cite{b19} employ different sensor measurements like cameras or LiDARs and Multimodal Large Language Models (MLLMs), which process sensor data for generating the final trajectory with real-time performance. Conversely, other works have designed and tested similar architectures in open-loop settings. EMMA\cite{b20}, OpenEMMA\cite{b21}, DriveVLM, VLM-AD\cite{b22}, DriveLLaVA\cite{b23} and ADAPT\cite{b24} exploit the broad knowledge of VLMs for scene understanding and trajectory generation, directly from sensor data.
These approaches have demonstrated significant efficacy as planning systems, eliminating the necessity for complex rule-based methodologies or elaborate cost functions by directly mapping environmental states to appropriate actions. However, the stringent safety-critical requirements inherent to autonomous driving systems raise substantial concerns regarding their operational reliability in authentic urban environments. A primary limitation stems from the susceptibility of VLMs to generate hallucinations when encountering novel scenarios. Furthermore, research conducted in DriveBench\cite{b10} has conclusively demonstrated that VLM behavior exhibits unpredictability, particularly when processing corrupted or degraded input data. This inconsistency suggests that model outputs are frequently not aligned with their intrinsic reasoning capabilities, but are instead influenced by inherent biases within training datasets and pattern recognition artifacts that can adversely affect the fine-tuning process and, consequently, compromise the integrity of their decision-making mechanisms.

\subsection{VLMs Integrated with End-to-End Systems}
To address these critical safety concerns while preserving the extensive world knowledge of Visual Language Models, researchers have developed sophisticated hybrid architectures that strategically integrate VLMs with end-to-end autonomous driving frameworks. These systems establish structured interoperability between VLMs' semantic reasoning capabilities and the operational efficiency of end-to-end driving systems, thereby mitigating individual limitations while enhancing collective performance in complex driving scenarios. Senna\cite{b2} employs a fine-tuned VLM as high-level decision-maker which supervises the trajectory generation process of the VAD\cite{b1} planning component. Hint-AD\cite{b25} integrates an end-to-end system like UniAD\cite{b26} with a VLM, which takes as input intermediate output tokens from perception, prediction and planning modules in order to provide scene descriptions and planning explanations aligned with internal states. To enhance end-to-end driving models, VLM-E2E\cite{b27} incorporates semantic text descriptions into the training process. By leveraging these descriptions, the model learns to focus its attention in a more semantically meaningful way, guided by visual-language models. 

These prior approaches exhibit notable limitations in their integration frameworks. While they successfully incorporate VLMs with end-to-end autonomous systems to provide either high-level planning supervision or explanatory capabilities, they consistently fail to unify both functionalities within a single architecture. This integration gap is particularly significant given that both planning supervision and explainability represent critical components for ensuring safety in autonomous driving deployments. 

The persistent absence of such comprehensive functionality in existing systems serves as the primary motivation for our research. Our work specifically addresses this limitation by developing an integrated framework that simultaneously provides robust planning supervision through VLM-guided decision-making while offering transparent, natural language explanations of driving behaviors, thus enhancing both operational safety and system trustworthiness within a unified architectural approach.

\section{VLAD Architecture}
\begin{figure*}[thpb]
  \centering
  \includegraphics[
    width=\textwidth]{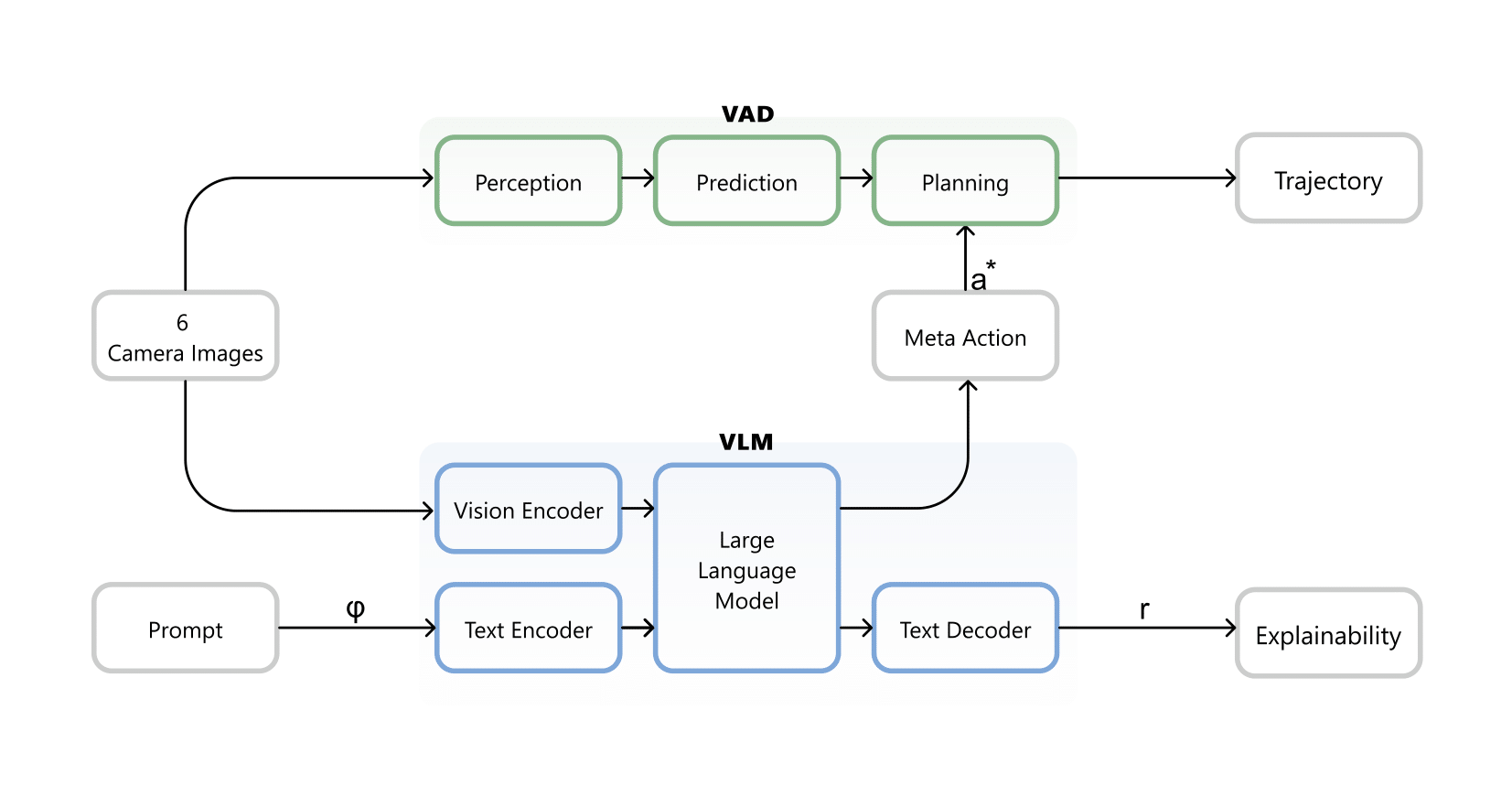}    
  \caption{\textbf{VLAD Architecture.} VLAD consists of two modules: a VLM integrated with an end-to-end system, VAD. The VLM encodes multi-view images and outputs a high-level command along with explanations of the selected behavior, based on the driving scenario. VAD processes the high-level command along with camera images and produces the final trajectory.}
  \label{VLAD-wide}
\end{figure*}

This section presents the detailed architecture of \textbf{VLAD}\footnote{Vision Language Autonomous Driving.}, our hybrid framework that integrates a Vision-and-Language Model (VLM) with an end-to-end driving pipeline. As illustrated in Fig.~\ref{VLAD-wide}, VLAD comprises two principal components:
\begin{enumerate}
    \item \textbf{Vectorized Autonomous Driving (VAD)}: a transformer-based end-to-end module based on\cite{b1} that takes as input multi-camera imagery and high-level navigational commands to produce a continuous trajectory.
    \item \textbf{Vision Language Model (VLM)}: a multimodal block built upon CLIP's ViT-L/14\cite{b28} backbone and a large language model (LLM) that reasons over scene context to output \emph{meta-actions} alongside textual justifications.
\end{enumerate}





\subsection{Vectorized Autonomous Driving (VAD) Module}
VAD is structured around three modules. \textbf{1.} The perception module detects moving objects and creates a local map. \textbf{2.} Following this, the motion prediction module estimates the future paths of these objects. \textbf{3.} The planning module then uses planning tokens, engaging with scene features via attention, to output the intended trajectory.
The driving scene is modeled using Transformer decoders over ego, agent, and map queries:

\paragraph{Ego–Agent Interaction}
\begin{equation}
\begin{split}
Q'_{\mathrm{ego}} & = \mathrm{TransDec}(q=Q_{\mathrm{ego}}, k=Q_a, v=Q_a,\\
\quad qpos & =PE_1(p_{\mathrm{ego}}), kpos=PE_1(p_a))
\end{split}
\end{equation}

\paragraph{Ego–Map Interaction}
\begin{equation}
\begin{split}
Q''_{\mathrm{ego}} & = \mathrm{TransDec}(q=Q'_{\mathrm{ego}}, k=Q_m, v=Q_m,\\
\quad qpos & =PE_2(p_{\mathrm{ego}}), kpos=PE_2(p_m))
\end{split}
\end{equation}

\paragraph{Planning Head}
\begin{equation}
\hat V_{\mathrm{ego}} = \mathrm{PlanHead}([Q'_{\mathrm{ego}},Q''_{\mathrm{ego}},s_{\mathrm{ego}}], c)
\end{equation}

Here, $Q_{\mathrm{ego}}$ is a learnable ego query, $Q_a$ and $Q_m$ are agent and map queries, $PE_1,PE_2$ are MLPs embedding positions $p$, $s_{\mathrm{ego}}$ the ego state and $c$ the high-level driving command. PlanHead decodes these into the future trajectory $\hat V_{\mathrm{ego}}\in\mathbb{R}^{T_f\times2}$, where $T_f$ denotes the number of future timestamps.

\subsection{Vision Language Model (VLM) Module}
The VLM block improves operational reliability by interpreting the scene in natural language and generating \emph{meta-actions}. It includes:
\begin{itemize}
    \item \textbf{Vision Encoder}: CLIP's ViT-L/14 \cite{b28}, which partitions each image into $P\times P$ patches, embeds them into tokens $z_i\in\mathbb{R}^{N\times D}$, and applies self-attention to produce global image features.
    \item \textbf{Large Language Model}: Vicuna-v1.5-7b \cite{b4}, an auto-regressive large language model, which receives concatenated visual tokens and a prompt template $\phi$ designed to query spatial relations (e.g., "Given the front camera image and agent positions, what is the recommended maneuver?").
\end{itemize}

Formally, let $Z=\{z_i\}_{i=1}^6$ be the sequence of visual embeddings and $\phi$ the tokenized prompt. The LLM cross-attends to $Z$ during each decoding step\cite{b29}:
\begin{equation}
\begin{split}
    & h_t = \mathrm{LM}\bigl(h_{t-1}, \mathrm{CrossAttn}(h_{t-1}, Z)\bigr),\\
   & \mathrm{CrossAttn}(Q,K,V) = \mathrm{softmax}\Bigl(\frac{QK^\top}{\sqrt{d_k}}\Bigr)V,
\end{split}
\end{equation}
where $h_t$ represents the hidden state at decoding step $t$, $d_k$ is the dimension of the key vectors, $Q'=W_Q'h_{t-1}$, $K'=W_K'Z$, and $V'=W_V'Z$ (being $W_Q', W_K', W_V'$ the learnable projection matrices). The decoder outputs:
\begin{enumerate}
    \item A discrete \emph{meta-action} label $a^*\in\mathcal{A}$ (e.g., \texttt{\small GO\_STRAIGHT}, \texttt{\small TURN\_LEFT}). $\mathcal{A}$ includes high-level commands along lateral directions.
    \item A textual rationale $r\in\Sigma^*$ explaining the decision.
\end{enumerate}
This dual output enables both direct control guidance and interpretability, crucial for safety validation and human oversight.

\begin{figure*}[thpb]
    \centering
    \includegraphics[width=\textwidth]{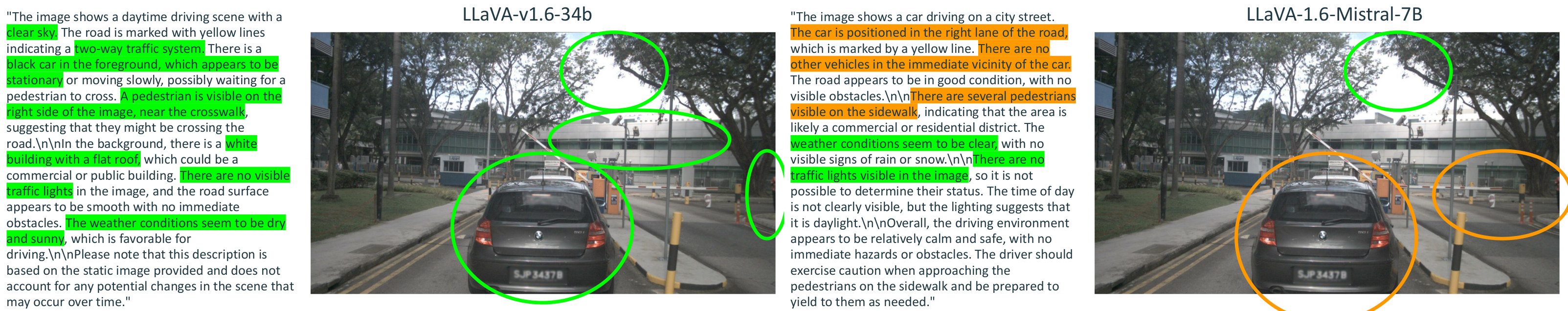}
    \caption{\textbf{Comparison of different LLaVA models.} The image shows a real-world sample data from nuScenes, where two models, namely LLaVA-v1.6-34B and LLaVA-1.6-Mistral-7B, are prompted to generate a scene description as driving agents. LLaVA-v1.6-34B is able to generate an accurate description, analyzing well the driving setting. Conversely, LLaVA-1.6-Mistral-7B, is not able to correctly detect important details like the car in the foreground and the lane in which the ego vehicle is driving. This highlights the importance of using a bigger model like LLaVA-v1.6-34B for generating the QA data. Text highlighted in green indicates correct alignment with the corresponding green circle in the image. Conversely, red highlighting signifies incorrect alignment with the corresponding red circles, often due to the model's inability to identify certain driving scene details.}
    \label{LLaVA comparison}
\end{figure*}

\subsection{Question-Answer Data Generation}
In order to enhance the VLM driving capabilities, we adopt a fine-tuning strategy on a custom Question-Answer (QA) Dataset, generated offline using LLaVA-v1.6-34b\cite{b3}, which due to its higher number of parameters, it is able to produce more precise outputs, at the cost of a much higher inference speed, as illustrated in Fig.~\ref{LLaVA comparison}.
The QA Dataset consists of question-answer pairs, designed to increase the perception, prediction and planning capabilities of the VLM in driving settings.

\subsubsection{Perception}
The autonomous driving system employs multi-modal perception by prompting the model with surround-view camera imagery alongside structured queries. These queries instruct the model to comprehensively describe the visual scenes captured by each camera, with particular emphasis on vulnerable road users (pedestrians, cyclists, and motorcyclists), traffic signal states, and other relevant static or dynamic actors within the environment. This directed attention mechanism ensures the model prioritizes safety-critical objects and conditions during the driving task.

\subsubsection{Prediction}
For each actor identified through the camera perception system, the model is tasked with generating trajectory predictions for future motion patterns. This predictive component facilitates the model's understanding of spatial-temporal relationships between dynamic objects and the surrounding environment, enabling anticipatory decision-making based on expected scene evolution.

\subsubsection{Planning}
In this final stage, given the comprehensive understanding of both current environmental states and predicted future conditions, the model generates a \emph{meta-action} that provides high-level guidance for the ego vehicle navigation, along with natural language explanations of the selected choice. Through this structured prompting approach, the model develops robust planning capabilities that prioritize safe driving behaviors in complex, dynamic environments.

These QA pairs are constructed using ground-truth observations and precise agent positions with their corresponding future trajectories, providing the model with accurate data for improved contextual understanding and decision-making. Following the data generation pipeline, we performed extensive fine-tuning of our Vision-Language Model (VLM) using a substantive dataset comprising 365,666 question-answer pairs. This comprehensive training procedure was complemented by a rigorous evaluation protocol conducted on a separate validation set containing 76,930 distinct question-answer pairs, ensuring proper assessment of the model's generalization capabilities across various driving scenarios and reasoning tasks.
Samples of this dataset are available on our GitHub repository\footnote{\url{https://github.com/1ASL-gifu/VLAD}}.

\subsection{Training Methodology}
VLAD framework is developed through a strategic two-stage training protocol. 
\subsubsection{First Stage}
We maintain the VAD component in a frozen state while comprehensively training the VLM on our curated Question-Answer Dataset. This specialized training enables the VLM to acquire domain-specific knowledge of driving environments and develop sophisticated planning capabilities for generating safe \emph{meta-actions}. This approach ensures the model establishes robust contextual understanding before implementation in trajectory generation.
\subsubsection{Second Stage}
We invert the training paradigm by freezing the trained VLM while conducting end-to-end training of the VAD component on the nuScenes dataset. This phase focuses on optimizing the system's ability to process multi-view camera imagery in conjunction with high-level driving commands, ultimately producing trajectories that demonstrate both human-like behavior and adherence to safety navigation. This sequential training methodology facilitates specialized learning at each stage while maximizing the complementary capabilities of both subsystems within the integrated VLAD architecture.

\section{Experiments and Results}
\subsection{Experimental Settings}
\subsubsection{Dataset}
The efficacy of the VLAD framework was rigorously evaluated across nuScenes dataset, which comprises 1,000 well curated driving scenarios, each with approximately 20 seconds of temporal duration. This dataset provides a rich multimodal sensory collection, including surround-view camera imagery, LiDAR point cloud data, and additional complementary sensor modalities. Furthermore, nuScenes offers extensive annotation resources, encompassing precisely labeled 3D bounding boxes, high-resolution ego vehicle trajectory information, and detailed semantic map representations of the driving environment. These annotations facilitate comprehensive assessment of perception, prediction, and planning capabilities within complex urban driving scenarios.

\subsubsection{Metrics}
In order to evaluate VLAD capabilities to plan a safe and human-like trajectory, we employ collision rate and displacement error metrics, as defined in\cite{b5}. Moreover, to assess its ability to provide accurate planning explanations, we adopt different metrics such as BLEU\cite{b30}, METEOR\cite{b31}, ROUGE-L\cite{b32}, CIDEr\cite{b33} and GPT-Score\cite{b34}. Additionally, we compute the accuracy of the model to provide correct \emph{meta-actions} based on the specific driving scenarios.
Finally, we conduct a comprehensive analysis of explanatory output inference times across multiple textual formats. We evaluate the model's performance with and without 4-bit quantization techniques to assess its viability for real-time applications.

\subsubsection{Implementation Details}
We explored two fine-tuning methodologies using DeepSpeed with ZeRO-3\cite{b40} parallelism: Low-Rank Adaptation (LoRA)\cite{b35} and full-parameter fine-tuning.

\textbf{Fine-tuning Approaches.}
Our LoRA implementation uses trainable low-rank matrices ($r=128$, $\alpha=256$), significantly reducing trainable parameters while maintaining performance. For full-parameter fine-tuning, we updated all weights in both language and vision components. Both approaches leveraged the pretrained CLIP ViT‐L/14 vision encoder (with its image‐adapter weights unfrozen and not re-initialized), taking the final hidden‐layer embeddings as visual features. These embeddings were then passed through a two-layer MLP projector (GELU activations) for multimodal fusion.

\textbf{Training Configuration.}
Both methodologies shared identical hyperparameters: learning rate of $2 \times 10^{-5}$, cosine schedule with 3\% warmup, no weight decay, per-device batch size of 8, gradient accumulation steps of 3, and bfloat16 precision. Models were trained for one epoch (an additional full-parameter fine-tuning was conducted for 10 epochs) with gradient checkpointing enabled and sequence length limited to 2048 tokens. Checkpoints were saved every 200 steps, with padding-based aspect ratio preservation for input images and modality-based sequence grouping for efficient processing.

\subsection{Main Results}
\subsubsection{VLAD Trajectory Planning}
We present in Table~\ref{tabt} the trajectory planning performance of VLAD, conducted on nuScenes dataset. VLAD outperforms other state-of-the-art systems in the collision rate metrics. Compared to VAD\cite{b1}, VLAD reduces the average collision rate by 31.82\%, and compared to Senna\cite{b2}, the baseline with the best performance, VLAD reduces the average collision rate by 16.7\%,  establishing a new benchmark in nuScenes dataset (Senna version trained on nuScenes and not fine-tuned on DriveX Dataset was evaluated for a fair comparison). Thanks to the world knowledge encapsulated in VLMs and our fine-tuning approach, VLAD is able to perceive dangerous situations and guide the ego vehicle with the safest driving behaviour.

\begin{table}[t]
\centering
\caption{Comparison of trajectory displacement error (L2) and collision rates over different horizons.}
\label{tabt}
\scriptsize  
\begin{tabular}{lcccccccc}
\toprule
\textbf{Method}
  & \multicolumn{4}{c}{\textbf{L2 (m)} $\downarrow$}
  & \multicolumn{4}{c}{\textbf{Collision (\%)} $\downarrow$} \\
\cmidrule(lr){2-5} \cmidrule(lr){6-9}
  & \textbf{1s} & \textbf{2s} & \textbf{3s} & \textbf{Avg.}
  & \textbf{1s} & \textbf{2s} & \textbf{3s} & \textbf{Avg.} \\
\midrule
ST-P3\cite{b36}        & 1.33 & 2.11 & 2.90 & 2.11 & 0.23 & 0.62 & 1.27 & 0.71 \\
Vanilla\cite{b37}       &  0.50   &  1.25   &  2.80   & 1.51 &  0.68   &  0.98   & 1.92  & 1.19    \\
NMP\cite{b38}    &  0.61   &  1.44   &  3.18   & 1.74 &  0.66   &  0.90   & 2.34  & 1.3    \\
FF\cite{b39}        & 0.56 & 1.27 & 3.08 & 1.64 & 0.65 & 0.86 & 1.64 & 1.05 \\
VLM-E2E\cite{b27}        & 1.22 & 1.94 & 2.68 & 1.95 & 0.26 & 0.60 & 1.17 & 0.68\\
\midrule
UniAD~\cite{b26}      & 0.48 & 0.96 & 1.65 & 1.03 & 0.05 & 0.17 & 0.71 & 0.31 \\
VAD-Tiny\cite{b1} & 0.46 & 0.76 & 1.12 & 0.78 & 0.21 & 0.35 & 0.58 & 0.38 \\
VAD-Base~\cite{b1}  & \textbf{0.41} & \textbf{0.70} & \textbf{1.05} & \textbf{0.72} & 0.07 & 0.17 & 0.41 & 0.22 \\
Senna\cite{b2} & \textbf{0.37} & \textbf{0.54} & \textbf{0.86} & \textbf{0.59} & 0.09 & 0.12 & 0.33 & 0.18 \\   
\midrule
VLAD*   & 0.65 & 1.09 & 1.60 & 1.11 & \textbf{0.05} & 0.23 & 0.60 & 0.29 \\
VLAD$^\dagger$               & 0.58 & 0.94 & 1.33 & 0.95 & \textbf{0.02} & 0.13 & 0.34 & \textbf{0.16} \\
VLAD$^\ddagger$     & 0.58 & 0.93 & 1.32 & 0.94 & \textbf{0.02} & \textbf{0.12} & \textbf{0.33} & \textbf{0.15} \\
\bottomrule
\multicolumn{8}{l}{$^{\mathrm{a}}$ VLAD*: LoRA fine-tuning for 1 epoch.} \\
\multicolumn{8}{l}{$^{\mathrm{b}}$ VLAD$^\dagger$: full-parameter fine-tuning for 1 epoch.} \\
\multicolumn{8}{l}{$^{\mathrm{c}}$ VLAD$^\ddagger$: full-parameter fine-tuning for 10 epochs.}
\end{tabular}
\end{table}

    \begin{figure}[thpb]
    \centering
    \framebox{\parbox{3.3in}{\includegraphics[width=\linewidth]{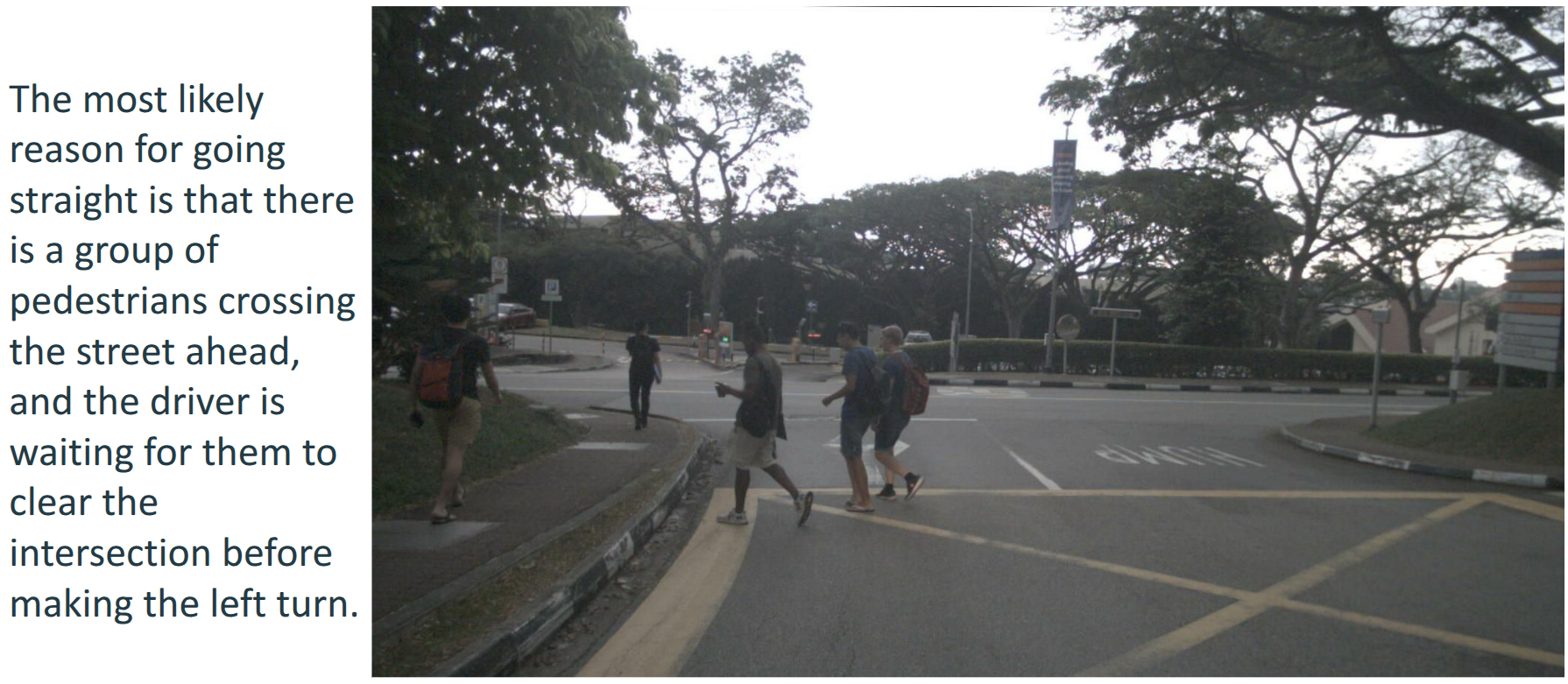}}}
    \caption{\textbf{VLAD Safety-First Approach Demonstration.} Despite an original high-level command to turn left, the VLM detects pedestrians crossing the intersection and keeps its recommendation to "proceed straight", postponing the execution of the turn until the road is clear, as also justified in its explanation. This adaptation enables the end-to-end system to focus on the vulnerable road users, reducing vehicle speed and preventing a collision. While this safety-oriented decision increases displacement error relative to the ground truth trajectory, it shows how our hybrid architecture effectively balances safety against path optimization by postponing the planned maneuver until safe conditions are established.}
    \label{ped}
\end{figure}

However, relative to the VAD and Senna baselines, our hybrid system incurs a slightly higher trajectory displacement error. Nonetheless, it still outperforms UniAD\cite{b26} and all other evaluated methods. This apparent contradiction between safety metrics and trajectory precision can be attributed to several key factors that characterize the fundamental nature of our hybrid architecture:

\begin{enumerate}[label=(\roman*)]
    \item The VLM demonstrates superior semantic understanding of driving scenes through its training on diverse question-answer pairs that specifically emphasized vulnerable road users and hazardous conditions. When processing camera inputs, the VLM appears to identify safety-critical elements that may not be equally weighted in the end-to-end system's internal representations. Consequently, the generated meta-actions occasionally prioritize collision avoidance behaviors that intentionally deviate from the most direct path when potential hazards are detected. We show an example in Fig.~\ref{ped}.


    \item Our hybrid system introduces a layer of semantic reasoning through the VLM's meta-action generation. This semantic layer evaluates scene context beyond pure geometry, occasionally recommending conservative decisions that prioritize keeping safe distances from other traffic participants. These safety-oriented decisions naturally manifest as slight deviations from the geometrically optimal paths represented in ground truth trajectories.
    \item The improved collision avoidance without corresponding trajectory precision suggests our hybrid system benefits from complementary perception strengths. The VLM's capacity to interpret complex interactions, predict pedestrian intentions, and understand unusual driving scenarios provides valuable high-level guidance, while the end-to-end system maintains expertise in control execution. The slight increase in displacement error thus represents a rational compromise wherein geometric optimality is occasionally sacrificed for enhanced operational safety.
    \item Analysis of specific scenarios revealed that at critical decision points where multiple viable paths exist, our hybrid system occasionally selected paths different from those chosen in ground truth demonstrations. These alternative paths, while equally valid and safer from a collision avoidance perspective, naturally accumulate displacement error relative to the reference trajectory. This observation suggests that displacement error alone may be an incomplete metric for evaluating autonomous driving systems that incorporate higher-level reasoning about scene safety.
    \item Our findings position the hybrid system at a different point on the safety-efficiency frontier compared to the baseline end-to-end approach. The substantial reduction in collision rates demonstrates that incorporating vision-language understanding enables more robust operation in complex environments, particularly those involving unpredictable road users or unusual scenarios not extensively represented in the training data. This safety advantage outweighs the marginal decrease in trajectory precision, especially for real-world deployment scenarios where collision avoidance represents the paramount objective.
\end{enumerate}

\subsubsection{VLAD Planning Explainability}
We now shift our focus on analyzing VLAD's ability to provide accurate natural language explanations that justify the selected high-level command, based on the driving scenario. This analysis highlights the importance on providing precise descriptions, increasing transparency and trustworthiness of the end-to-end architecture.
We compare VLAD explanations with the ground-truth ones generated offline in our QA Dataset, and different metrics are employed.
Table~\ref{met} reveals significant performance differences across fine-tuning approaches for our VLAD system. The baseline without fine-tuning and LoRA fine-tuning for one epoch (VLAD and VLAD*) demonstrate comparable performance, with minimal differences across all evaluation metrics. This suggests that parameter-efficient fine-tuning via LoRA provides insufficient adaptation for the autonomous driving domain when limited to a single training epoch.
In contrast, full-parameter fine-tuning (VLAD$^\dagger$) produces dramatic improvements, with BLEU scores increasing from 19.83 to 64.60 (3.2×), and CIDEr scores improving from 0.11 to 3.71 (33.7×). This substantial performance gap indicates that the complex multimodal reasoning required for autonomous driving necessitates comprehensive adaptation throughout the model's parameter space, which low-rank adaptation techniques cannot provide.
Extending full-parameter fine-tuning to ten epochs (VLAD$^\ddagger$) yields only marginal additional gains, suggesting that one epoch captures most of the adaptation benefits. These findings highlight a critical trade-off: while parameter-efficient methods offer computational advantages, the specialized nature of autonomous driving tasks—interpreting multiple camera inputs and generating appropriate driving-related descriptions—requires more extensive parameter adjustments than LoRA can provide within limited training regimes.

This pattern also arises when evaluating the planning efficacy of various VLAD configurations in generating appropriate \emph{meta-actions}, as presented in Table~\ref{plan}. The baseline model and its LoRA fine-tuned counterpart demonstrate comparable performance metrics, whereas the full-parameter fine-tuned variants exhibit substantial improvements, effectively doubling the planning accuracy. These consistent findings across multiple evaluation dimensions underscore the critical importance of selecting appropriate fine-tuning methodologies when adapting foundation models to specialized domains such as autonomous driving, where domain-specific reasoning and multimodal integration capabilities are paramount.

\begin{table}[t]
\centering
\caption{Explanation‐quality metrics for VLAD variants.}
\label{met}
\scriptsize  
\begin{tabular}{lccccc}
\toprule
\textbf{Method}
  & {\textbf{BLEU} $\uparrow$}
  & {\textbf{METEOR} $\uparrow$} 
  & {\textbf{ROUGE-L} $\uparrow$} 
  & {\textbf{CIDEr} $\uparrow$}
  & {\textbf{GPT-Score} $\uparrow$} \\
\midrule

VLAD        & 19.96 & 37.84 & 39.28 & 0.17 & 1.86 \\
VLAD*       &  19.83   &  37.88   &  39.65   & 0.11 &  2.04    \\
VLAD$^\dagger$       & \textbf{64.60} & \textbf{73.27} & \textbf{72.40} & \textbf{3.71} & \textbf{3.76} \\
VLAD$^\ddagger$     & \textbf{64.90} & \textbf{73.48} & \textbf{72.97} & \textbf{3.82} & \textbf{3.78} \\
\bottomrule
\multicolumn{5}{l}{$^{\mathrm{a}}$ VLAD: without fine-tuning} \\
\multicolumn{5}{l}{$^{\mathrm{b}}$ VLAD*: LoRA fine-tuning for 1 epoch.} \\
\multicolumn{5}{l}{$^{\mathrm{c}}$ VLAD$^\dagger$: full-parameter fine-tuning for 1 epoch.} \\
\multicolumn{5}{l}{$^{\mathrm{d}}$ VLAD$^\ddagger$: full-parameter fine-tuning for 10 epochs.}
\end{tabular}
\end{table}

\begin{table}[t]
\centering
\caption{Planning accuracy.}
\label{plan}
\scriptsize  
\begin{tabular}{lccccc}
\toprule
\textbf{Method}
  & {\textbf{Accuracy (\%)} $\uparrow$} \\
\midrule
VLAD        &  44.15\\
VLAD*       &  44.36\\
VLAD$^\dagger$       & \textbf{90.15}  \\
VLAD$^\ddagger$     & \textbf{90.76} \\
\bottomrule
\end{tabular}
\end{table}

\begin{figure*}[!b] 
    \centering
    \includegraphics[width=\textwidth]{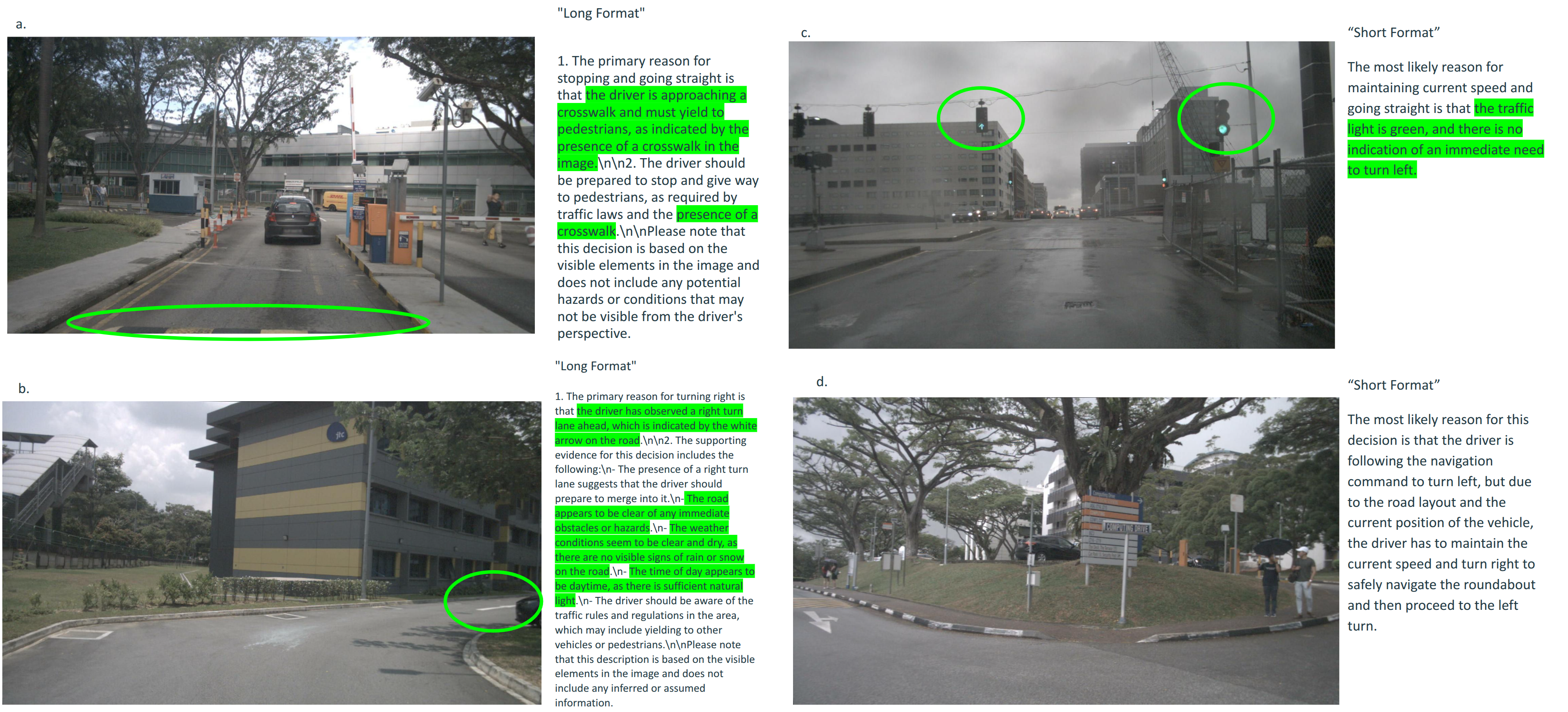}
    \caption{\textbf{Qualitative Results.} We present a comparative analysis of our model's explanatory capabilities in both formats. Panels \textbf{a} and \textbf{b} demonstrate long-format explanations, while panels \textbf{c} and \textbf{d} illustrate short-format examples.
The long-format examples show how the model identifies critical driving elements, including infrastructure (crosswalks, directional arrows) while providing comprehensive environmental descriptions. These detailed explanations reveal the model's perception processes.
Short-format examples deliver concise justifications for planning commands while maintaining informational integrity. Panel \textbf{d} showcases sophisticated reasoning: despite a high-level left turn command, the VLM recognizes the need for an initial right turn to navigate a roundabout before executing the left turn—demonstrating advanced spatial awareness and planning capabilities. This highlights explainable AI's essential role in autonomous driving systems, providing decision-making transparency while demonstrating the fine-tuned VLM's ability to interpret complex scenarios and generate safety-oriented commands aligned with real-world navigation requirements.}
    \label{dio}
\end{figure*}

\subsubsection{Real-Time Performance}

An essential consideration for Visual Language Models (VLMs) in autonomous driving applications is their ability to generate outputs—such as high-level commands or explanations—at frequencies compatible with real-time operation requirements.
To evaluate this aspect, we conducted a comprehensive analysis of inference time for our hybrid model, with results summarized in Table~\ref{inf}. Our investigation examined four distinct experimental configurations: long-format descriptions and short-format descriptions, both with and without 4-bit quantization.
In the long-format configuration, the model generates a detailed explanation of the selected high-level command, accompanied by environmental context and identification of relevant actors requiring caution. The short-format configuration, conversely, prompts the model to provide a concise justification for selecting a specific \emph{meta-action} within the given driving scenario.
While the short-format explanation offers less comprehensive insight than its longer counterpart, it successfully delivers essential justification and environmental assessment by focusing on critical decision factors. This approach more closely resembles human cognitive processes during driving, where rapid decision-making predominates over extended deliberation.
Additionally, we applied 4-bit quantization to both configurations to investigate potential inference time improvements while carefully monitoring any impact on accuracy. As demonstrated in Table~\ref{inf}, the long-format explanations—with and without quantization—exhibited inference times of approximately 3 seconds, which falls short of real-time performance requirements.
In contrast, the short-format explanations achieved remarkable inference times of approximately 0.8 seconds, making them viable candidates for real-time applications. Although this frequency may not satisfy the demands of certain autonomous driving algorithms or when generating meta-actions, it proves sufficient for planning explanations. This aligns with human cognitive limitations, as drivers cannot typically articulate justifications for their behaviors at higher frequencies, nor can humans process explanations delivered at such rates.
Our results indicate that while 4-bit quantization marginally reduces inference time in both configurations, the improvement is negligible, suggesting an unfavorable trade-off between processing speed and model accuracy. In Fig.~\ref{dio} we present some qualitative results of both long and short format explanations.

\begin{table}[t]
\centering
\caption{Inference Time Ablation Study}
\label{inf}
\scriptsize  
\begin{tabular}{lccccc}
\toprule
\textbf{Format}
  & {\textbf{Inference Time (s)} $\downarrow$} \\
\midrule
Long Format        &  3.407\\
Long Format (4-bit Q.)      &  3.332\\
Short Format       & \textbf{0.878}  \\
Short Format (4-bit Q.)    & \textbf{0.817} \\
\bottomrule
\end{tabular}
\end{table}

\section{Conclusions}
This paper presented VLAD (Vision-Language Autonomous Driving), a novel hybrid architecture that integrates a fine-tuned Visual Language Model (VLM) with VAD, a state-of-the-art end-to-end autonomous driving system. Through our specialized question-answer dataset, the VLM component demonstrated enhanced spatial awareness and significantly improved capabilities across perception, prediction, and planning domains specific to autonomous driving scenarios.
VLAD achieved comprehensive functionality by generating both trajectories and natural language explanations from camera imagery alone, without requiring additional sensor modalities. Our experimental evaluation on the nuScenes benchmark demonstrated that VLAD outperforms current state-of-the-art baseline systems, particularly in safety-critical metrics. Most notably, VLAD reduces collision rates by approximately 30\%, positioning our approach at the forefront of safe planning methodologies for autonomous vehicles.
We have further evaluated VLAD's explanatory capabilities through targeted quantitative metrics, demonstrating strong performance when compared to ground truth models. Qualitative analysis confirms VLAD's capacity for complex reasoning across diverse and challenging driving scenarios. Moreover, our system delivers concise natural language explanations at frequencies compatible with real-time operational requirements (0.8 seconds per explanation), addressing a critical need in explainable autonomous driving systems.
Future research directions include comprehensive closed-loop evaluations to assess VLAD's effectiveness in continuously guiding vehicles through dynamic environments. Additionally, we plan to develop aligned interpretability frameworks by leveraging intermediate representations from VAD components, thereby enhancing the precision and relevance of the system's explanatory outputs. Finally, we plan to test VLAD on Drivebench\cite{b10}, to quantitatively assess its explainability performance against other baselines. These developments will further advance the integration of explainable AI principles into autonomous driving technologies, addressing both technical performance requirements and human-centered trust considerations.

\section*{Acknowledgment}
We gratefully acknowledge the contributions of the graduate research assistants at the Intelligent Autonomous Systems Laboratory (iASL) for their technical support in configuring and maintaining the computational infrastructure essential for our experimental evaluations. We extend our sincere appreciation to DENSO CORPORATION for their generous support and provision of computational resources that made possible the training and validation of our large-scale models. The authors thank Chiara Corti for her graphic design work on Fig.~\ref{VLAD-wide}. This research represents a collaborative effort between academia and industry partners committed to advancing the field of autonomous driving technologies.


\begin{thebibliography}{00}
\bibitem{b1} Bo Jiang, Shaoyu Chen, Qing Xu, ``VAD: Vectorized Scene Representation for Efficient Autonomous Driving,'' ICCV, 2023.
\bibitem{b2} Bo Jiang, Shaoyu Chen, Bencheng Liao, ``Senna: Bridging Large Vision-Language Models and End-to-End Autonomous Driving,`` arXiv, 2024.
\bibitem{b3} Haotian Liu, Chunyuan Li, Qingyang Wu, ``Visual Instruction Tuning,`` NeurIPS, 2023.
\bibitem{b4} Lianmin Zheng, Wei-Lin Chiang, Ying Sheng, ``Judging LLM-as-a-Judge with MT-Bench and Chatbot Arena,`` NeurIPS, 2023.
\bibitem{b5} Holger Caesar, Varun Bankiti, Alex H. Lang, ``nuScenes: A multimodal dataset for autonomous driving,'' CVPR, 2020.
\bibitem{b6} Hao Sha, Yao Mu, Yuxuan Jiang, ``LanguageMPC: Large Language Models as Decision Makers for Autonomous Driving,'' arXiv, 2025.
\bibitem{b7} Zhenhua Xu, Yujia Zhang, Enze Xie, ``DriveGPT4: Interpretable End-to-end Autonomous Driving via Large Language Model,'' IEEE ROBOTICS AND AUTOMATION LETTERS, 2024.
\bibitem{b8} Ana-Maria Marcu, Long Chen, Jan Hünermann, ``LingoQA: Visual Question Answering for
Autonomous Driving,'' ECCV, 2024.
\bibitem{b9} Chonghao Sima, Katrin Renz, Kashyap Chitta, ``DriveLM: Driving with Graph Visual Question Answering,`` ECCV, 2024.
\bibitem{b10} Shaoyuan Xie, Lingdong Kong, Yuhao Dong, ``Are VLMs Ready for Autonomous Driving? An Empirical Study from the Reliability, Data, and Metric Perspectives,`` arXiv, 2025.
\bibitem{b11} Alexey Dosovitskiy, German Ros, Felipe Codevilla, ``CARLA: An Open Urban Driving Simulator,`` CoRL, 2017.
\bibitem{b12} Tsun-Hsuan Wang, Alaa Maalouf, Wei Xiao, ``Drive Anywhere: Generalizable End-to-end Autonomous Driving with Multi-modal Foundation Models,`` arXiv, 2024.
\bibitem{b13} Hao Shao, Yuxuan Hu, Letian Wang, ``LMDrive: Closed-Loop End-to-End Driving with Large Language Models,`` arXiv, 2024.
\bibitem{b14} Katrin Renz, Long Chen, Ana-Maria Marcu, ``CarLLaVA: Vision language models for camera-only closed-loop driving,`` arXiv, 2024.
\bibitem{b15} Wenhai Wang, Jiangwei Xie, ChuanYang Hu, ``DriveMLM: Aligning Multi-Modal Large Language Models with Behavioral Planning States for Autonomous Driving,`` arXiv, 2023.
\bibitem{b16} Xiaoyu Tian, Junru Gu, Bailin Li, ``DriveVLM: The Convergence of Autonomous Driving and Large Vision-Language Models,`` CoRL, 2024.
\bibitem{b17} Zilin Huang, Zihao Sheng, Yansong Qu, ``VLM-RL: A Unified Vision Language Models and Reinforcement Learning Framework for Safe Autonomous Driving,`` arXiv, 2024.
\bibitem{b18} Katrin Renz, Long Chen, Elahe Arani, ``SimLingo: Vision-Only Closed-Loop Autonomous Driving with Language-Action Alignment,`` CVPR, 2025.
\bibitem{b19} Xiangru Mu, Tong Qin, Songan Zhang, ``Pix2Planning: End-to-End Planning by Vision-language Model for Autonomous Driving on Carla Simulator,`` IEEE Intelligent Vehicles Symposium (IV), 2024.
\bibitem{b20} Jyh-Jing Hwang, Runsheng Xu, Hubert Lin, ``EMMA: End-to-End Multimodal Model for Autonomous Driving,`` arXiv, 2024.
\bibitem{b21} Shuo Xing, Chengyuan Qian, Yuping Wang, ``OpenEMMA: Open-Source Multimodal Model for End-to-End Autonomous Driving,`` LLVM-AD, 2025.
\bibitem{b22} Yi Xu, Yuxin Hu, Zaiwei Zhang, ``VLM-AD: End-to-End Autonomous Driving through Vision-Language Model Supervision,`` arXiv, 2024.
\bibitem{b23} Rui Zhao, Qirui Yuan, Jinyu Li, ``DriveLLaVA: Human-Level Behavior Decisions via Vision Language Model,`` Sensors (Basel), 2024.
\bibitem{b24} Bu Jin, Xinyu Liu, Yupeng Zheng, ``ADAPT: Action-aware Driving Caption Transformer,`` ICRA, 2023.
\bibitem{b25} Ding, Kairui, Chen, ``Hint-AD: Holistically Aligned Interpretability in End-to-End Autonomous Driving,`` CoRL, 2024.
\bibitem{b26} Yihan Hu, Jiazhi Yang, Li Chen, ``Planning-Oriented Autonomous Driving,`` CVPR, 2023.
\bibitem{b27} Pei Liu, Haipeng Liu, Haichao Liu, ``VLM-E2E: Enhancing End-to-End Autonomous Driving with Multimodal Driver Attention Fusion,`` arXiv, 2025.
\bibitem{b28} Alec Radford, Jong Wook Kim, Chris Hallacy, ``Learning Transferable Visual Models From Natural Language Supervision,`` International Conference on Machine Learning, 2021.
\bibitem{b29} Ashish Vaswani, Noam Shazeer, Niki Parmar, ``Attention Is All You Need,`` NeurIPS, 2017.
\bibitem{b30} K. Papineni, S. Roukos, T. Ward, ``Bleu: a method for automatic evaluation of machine translation,`` ACL, 2002.
\bibitem{b31} S. Banerjee and A. Lavie, ``Meteor: An automatic metric for mt evaluation
with improved correlation with human judgments,`` ACL, 2005.
\bibitem{b32} Chin-Yew Lin, ``ROUGE: A Package for Automatic Evaluation of Summaries,`` Text summarization branches, 2004.
\bibitem{b33} R. Vedantam, C. Lawrence Zitnick, and D. Parikh, ``Cider: Consensusbased image description evaluation,`` CVPR, 2015.
\bibitem{b34} Jinlan Fu, See-Kiong Ng, Zhengbao Jiang, ``GPTScore: Evaluate as You Desire,`` arXiv, 2023.
\bibitem{b35} Edward J. Hu, Yelong Shen, Phillip Wallis, ``LoRA: Low-Rank Adaptation of Large Language Models,`` ICLR, 2022.
\bibitem{b36} S. Hu, L. Chen, P. Wu, ``St-p3: End-to-end
vision-based autonomous driving via spatial-temporal feature learning,`` ECCV, 2022.
\bibitem{b37} F. Codevilla, E. Santana, A. M. Lopez, ``Exploring the Limitations of Behavior Cloning for Autonomous Driving,`` ICCV, 2019.
\bibitem{b38} D. Chen, B. Zhou, V. Koltun, ``Learning by Cheating,`` CoRL, 2020.
\bibitem{b39} Peiyun Hu, Aaron Huang, John Dolan, ``Safe Local Motion Planning with Self-Supervised Freespace Forecasting,`` CVPR, 2021.
\bibitem{b40} Samyam Rajbhandari, Jeff Rasley, Olatunji Ruwase, ``ZeRO: Memory Optimizations Toward Training Trillion Parameter Models,`` SC20, 2020.
\end{thebibliography}
\end{document}